\documentclass[letterpaper, 10 pt, conference]{ieeeconf}  %

\IEEEoverridecommandlockouts                              %

\title{\LARGE \bf
Reconstructive Latent-Space Neural Radiance Fields for Efficient 3D Scene Representations 
}

\author{
Tristan Aumentado-Armstrong$^\text{1,2,4*}$, 
Ashkan Mirzaei$^\text{1,2*}$, 
Marcus A. Brubaker$^\text{1,3,4}$,
Jonathan Kelly$^\text{2,4}$,\\ 
Alex Levinshtein$^\text{1}$, 
Konstantinos G. Derpanis$^\text{1,3,4}$,
Igor Gilitschenski$^\text{2,4}$\\ 
\thanks{
$^\text{1}$Samsung AI Centre Toronto~~$^\text{2}$University of Toronto~~$^\text{3}$York University~~$^\text{4}$Vector Institute for AI.
$^\text{*}$Equal Contribution. Emails:
    {\tt 
     \{a.mirzaei, tristan.a\}@partner.samsung.com, \{jkelly, gilitschenski\}@cs.toronto.edu,
     \{kosta, mab\}@eecs.yorku.ca, alex.lev@samsung.com
    }
  }
}

\usepackage[T1]{fontenc}
\usepackage[symbol]{footmisc}

\usepackage{graphicx}
\usepackage{amsmath}
\usepackage{amssymb}
\usepackage{multirow}
\usepackage{makecell}

\makeatletter
\let\NAT@parse\undefined
\makeatother

\usepackage[pagebackref=true,breaklinks=true,colorlinks,bookmarks=false,citecolor=blue]{hyperref} %
\newcommand{\whi}{\widehat{I}}

\begin{document}

\maketitle
\thispagestyle{plain}
\pagestyle{plain}

\begin{abstract}
Neural Radiance Fields (NeRFs) have proven to be powerful 3D representations, capable of high quality novel view synthesis of complex scenes. While NeRFs have been applied to graphics, vision, and robotics, problems with slow rendering speed and characteristic visual artifacts prevent adoption in many use cases. In this work, we investigate combining an autoencoder (AE) with a NeRF, in which latent features (instead of colours) are rendered and then convolutionally decoded. The resulting latent-space NeRF can produce novel views with higher quality than standard colour-space NeRFs, as the AE can correct certain visual artifacts, while rendering over three times faster. Our work is orthogonal to other techniques for improving NeRF efficiency. Further, we can control the tradeoff between efficiency and image quality by shrinking the AE architecture, achieving over 13 times faster rendering with only a small drop in performance. We hope that our approach can form the basis of an efficient, yet high-fidelity, 3D scene representation for downstream tasks, especially when retaining differentiability is useful, as in many robotics scenarios requiring continual learning.
\end{abstract}

\section{Introduction}

Neural rendering techniques~\cite{advances.in.neural.rendering}  
    continue to grow in importance,
    particularly Neural Radiance Fields~\cite{original.nerf} (NeRFs), 
    which achieve state-of-the-art performance  
    in novel view synthesis and 3D-from-2D reconstruction. %
As a result, NeRFs have been utilized for a variety of applications, 
    not only in content 
        creation~\cite{in2n,arf,spinnerf,reference.guided.nerf},
    but also for many robotics tasks, including
    6-DoF tracking~\cite{robot2}, 
    pose estimation~\cite{robot3}, 
    surface recognition~\cite{robot4} or reconstruction~\cite{robot1}, 
    motion planning~\cite{robot5,robot6,robot7}, 
    reinforcement learning~\cite{driess2022reinforcement,shim2023snerl}, 
    tactile sensing~\cite{zhong2023touching}, and 
    data-driven simulation~\cite{block.nerf,yang2023reconstructing}. 
However, slow rendering and the qualitative artifacts of NeRFs impede further use cases in production.

\begin{figure}[t]
  \centering
   \includegraphics[width=0.99\linewidth]{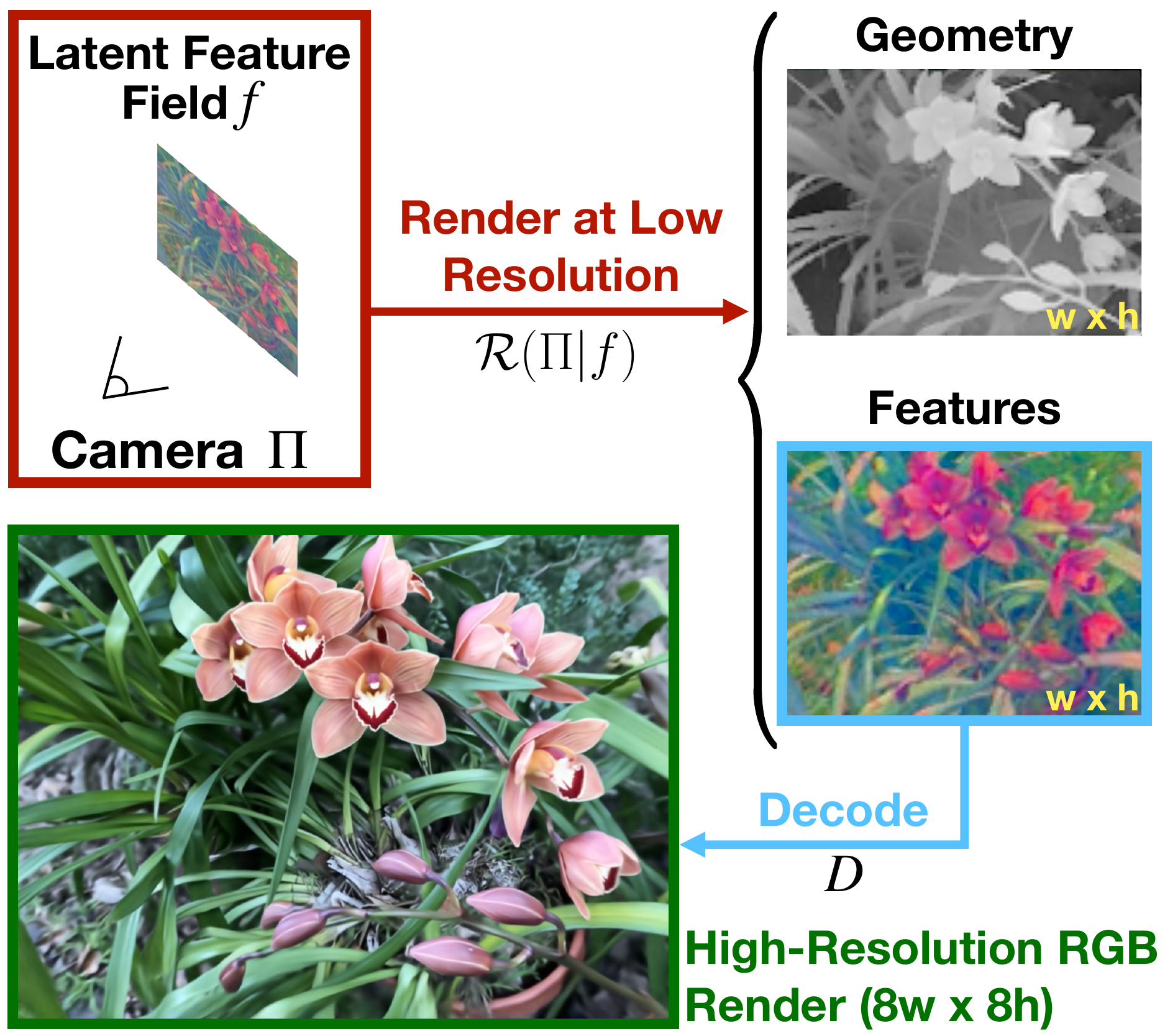}
   \caption{
        Illustration of the Reconstructive Latent-Space NeRF (ReLS-NeRF) rendering pipeline 
        (see \S\ref{sec:methods}).
        As shown in the upper-left inset,
        given a trained feature field, $f$, and a camera, $\Pi$, we can render a latent feature map at a \textit{low resolution}, as shown in the right inset. 
        The geometry of the scene, 
        encapsulated by the density field, 
        which determines the 3D structure of the feature render, is learned via an RGB component (as in a regular NeRF).
        A decoder, $D$, can then map the low-resolution feature maps to a high-resolution colour image
        (lower-left inset).
        We may view this process, which maps camera parameters to images, %
        as a form of neural rendering.
   }
   \label{fig:teaser}
\end{figure}

To render a single pixel, one major bottleneck is the need for multiple forward passes of a multilayer perceptron (MLP). 
Replacing or augmenting the MLP with alternative representations (e.g., voxel grids~\cite{plenoxels} or feature hash-tables~\cite{instant.ngp}) has been used to improve both training and inference speed. 
Baking NeRFs into other primitive representations has also been a popular approach~\cite{hedman2021baking,mobile.nerf,merf} for faster rendering. 
To reduce artifacts (e.g., ``floaters'' \cite{warburg2023nerfbusters}), 
different sampling methods~\cite{nerfstudio,barron2023zip,mipnerf,mipnerf.360}, 
radiance models~\cite{refnerf}, and 
scene contraction functions~\cite{nerfpp,mipnerf.360} 
have been proposed. 
Despite these advancements, 
    NeRFs still suffer from visual flaws and low rendering frame-rates. 

Importantly, such issues hamper the use of NeRFs for downstream tasks, 
If rendering is too slow, agents will be unable to apply NeRFs as an internal 3D representation of the scene.
Further, the solutions considered (often aimed at applications in computer graphics, for instance) may not be compatible with the requirements of other tasks.
For example, %
    meshification \cite{mobile.nerf,wan2023learning} enables fast rendering,
    but makes further online learning of the geometry significantly more difficult, 
    due to topological constraints \cite{pan2019deep} and additional optimization complexity (e.g., to handle self-intersections and other unnatural geometries) \cite{kato2018neural,wang2018pixel2mesh}.
We also do not wish to sacrifice too much representational fidelity (e.g., not including view-dependent effects \cite{sucar2021imap}) for speed, %
as less accurate visual output can limit downstream opportunities for  scene analysis. %
We therefore require a model that is capable of fast rendering and intra-task optimization (i.e., learning during an ongoing task), without sacrificing visual quality. 

In this paper, we propose an approach for solving these challenges that is orthogonal to existing methods.
By leveraging convolutional autoencoders (AEs), 
    we can define a ``NeRF'' operating in latent feature space 
    (rather than colour space),
    such that \textit{low}-resolution \textit{latent} renders can be decoded to \textit{high}-resolution RGB renders
    (see Fig.~\ref{fig:teaser}).
This offloads expensive MLP-based rendering computations to the low-cost AE, greatly improving efficiency.
Thus, we extend the standard NeRF architecture 
    to return point-wise latent vectors, 
    in addition to densities and colours (the latter used only in training). 
Since the decoder is simply another differentiable neural network, the ability to optimize the underlying 3D NeRF field is largely unchanged.
As it is used for scene reconstruction,
    we denote the resulting combined field
    a \textit{Reconstructive Latent-Space NeRF} (ReLS-NeRF).
Beyond improving rendering speed, 
    the AE can also act as an image prior, 
    fixing some of the artifacts associated with direct NeRF renders,
    and actually \textit{improving} representational fidelity. 
However, we also observe that the use of the AE in 
    ReLS-NeRF can introduce unique \textit{temporal} artifacts,  
    which existing image and video do not capture;
    hence, we define a novel metric that takes advantage of the geometric structure of the NeRF to detect them.

Overall, by fine-tuning a powerful pretrained AE,
    our model is able to render views several times faster, 
    while empirically improving in multiple image and video quality metrics.   
Further, we demonstrate a tradeoff between visual quality and rendering efficiency:
    by reducing the AE size, 
    we obtain a 13-fold speed-up, %
    with only a minor drop in quality.
In summary, we contribute 
(i) a novel approach to reconstructive 3D scene representation, via a latent-space NeRF that both improves rendering efficiency and outperforms existing work on standard image and video quality metrics;
(ii) a new evaluation metric, designed to detect temporal artifacts due to view inconsistencies, which existing metrics do not appear to capture; and
(iii) the ability to trade-off image quality and rendering speed via varying the AE architecture.

\section{Related Work}

\subsection{Improving NeRF efficiency}
While NeRFs produce results of extraordinary quality, 
    the speed of fitting (training) and rendering (inference)
    remains a bottleneck for adoption in a variety of applications 
    (e.g., \cite{mipnerf.360,block.nerf,turki2022mega}).
This has prompted a myriad of approaches to increasing their efficiency.
Feature grid architectures have proven effective
    in expediting fitting convergence
(e.g., \cite{wang2023f,sun2022direct,sun2022improved,barron2023zip,chen2022tensorf,chen2023factor,plenoxels,instant.ngp}).
Other approaches include utilizing depth \cite{ds.nerf}, better initializations \cite{tancik2021learned}, and 
pretraining conditional fields
(e.g., \cite{yu2021pixelnerf,wang2021ibrnet,johari2022geonerf}).
Such improvements can be readily utilized in our own framework.
Similarly, a number of methods have been proposed to enhance the efficiency of the volume rendering operation, which relies on an expensive Monte Carlo integration involving many independent neural network calls per pixel.
These include architectural modifications
\cite{garbin2021fastnerf,wadhwani2022squeezenerf,reiser2021kilonerf,kurz2022adanerf,kerbl3Dgaussians}, 
spatial acceleration structures \cite{yu2021plenoctrees}, 
``baking'' (precomputing and storing network outputs) \cite{hedman2021baking,merf}, 
improved sampling strategies \cite{piala2021terminerf,fang2021neusample,neff2021donerf,lin2022efficient,kondo2021vaxnerf}, or 
altering the integration method itself \cite{lindell2021autoint,wu2022diver}. 
Finally, several works eschew volume rendering itself. 
A number of representations \cite{sitzmann2021light,smith2023unsupervised,yenamandra2022fire,feng2022prif,aumentado2022representing,houchens2022neuralodf}
use only a single sample per pixel,
but struggle with geometric consistency and scalability.
Similarly, one can move to a mesh-based representation and use rasterization instead \cite{mobile.nerf,guo2023vmesh,wan2023learning}; 
however, this loses certain properties, 
such as amenability to further optimization or differentiable neural editing.
Though our approach improves rendering efficiency, 
    it is orthogonal to these methods,
    as it reduces the number of MLP calls per image
    by changing the output space of the NeRF itself.

\subsection{Feature-space NeRFs}
Other models have utilized \textit{neural feature fields} (NFFs), as opposed to ``radiance'' fields, 
where rendering is altered to output learned features instead.
Some NFFs \cite{tschernezki2022neural,kobayashi2022decomposing}
    learn to produce the outputs of pretrained 2D feature extractors;
similarly, several works have considered the use of 
    language-related features \cite{kerr2023lerf,blomqvist2023neural,shafiullah2022clip}
    and other segmentation signals
    \cite{zhi2021ilabel,zhi2021place,mirzaei2022laterf,spinnerf}
     to embed semantics into the NFF.
More closely related to our work are 
    generative modelling
    NFFs that decode rendered features into images
    via 
    generative adversarial networks \cite{gu2021stylenerf,niemeyer2021giraffe,xue2022giraffe}
    or diffusion models \cite{metzer2023latent,seo2023ditto,chan2023generative}.
In contrast, this paper considers the scene reconstruction problem, 
    using a latent representation potentially amenable to downstream tasks,
    and investigates issues related to view consistency.
In particular, the artifacts of generative methods are similar to those detected by our novel quality metric (namely, appearance inconsistencies across close frames or camera viewpoints; e.g., see  \cite{niemeyer2021giraffe}).

\begin{figure*}[t]
  \centering
   \includegraphics[width=1.0\linewidth]{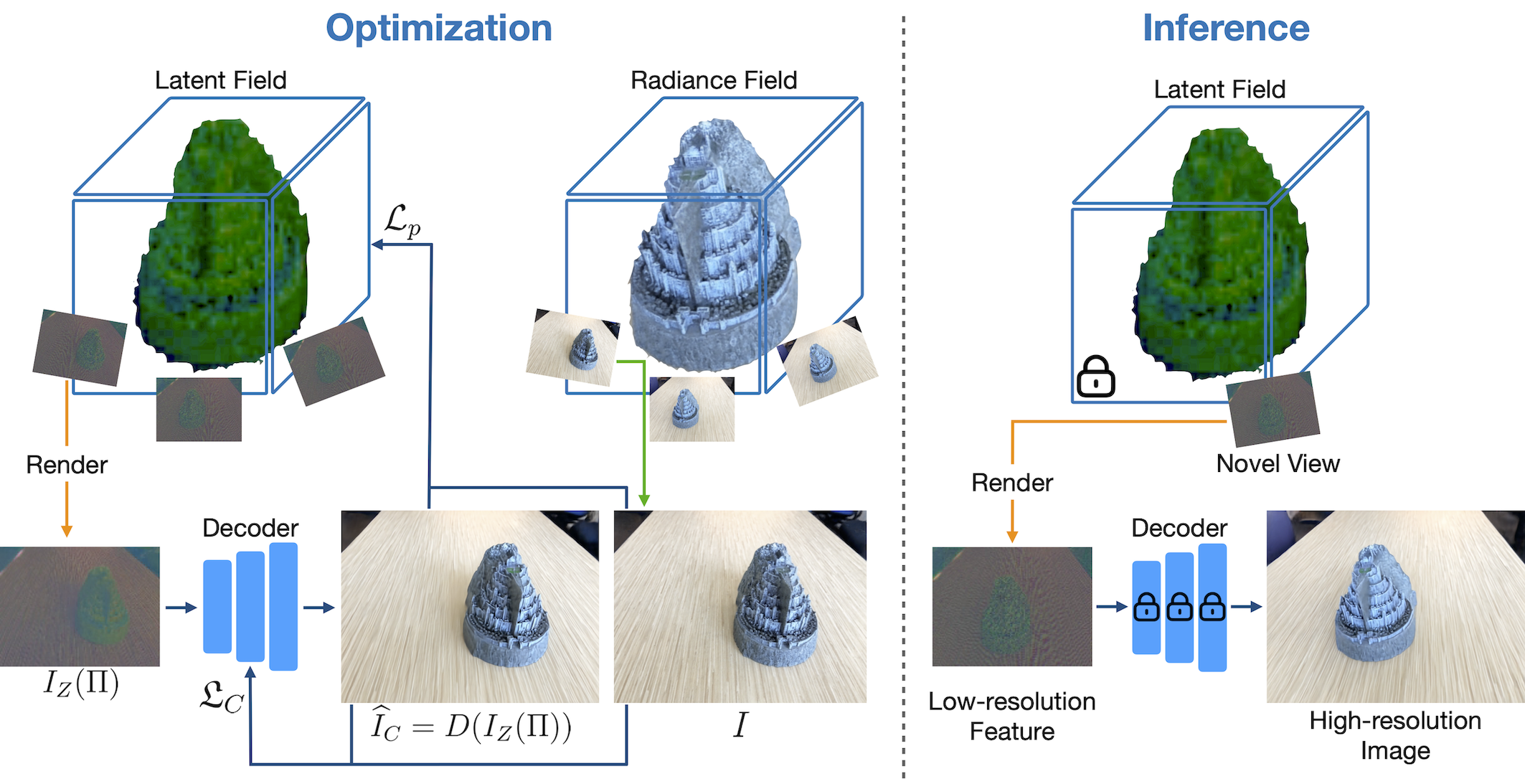}
   \caption{
    An overview of the ReLS-NeRF fitting and inference processes.
    \textit{Left: optimization approach}.
    The radiance (colour) field is fit to RGB captures, as in the standard NeRF~\cite{original.nerf}. 
    Given camera parameters, $\Pi$, ReLS-NeRF renders feature maps in the latent $Z$-space
        defined by a convolutional autoencoder (AE), $D \circ E$,
        for which arbitrary views can be decoded into image space. 
    The discrepancy between the decoded features and the corresponding images 
    (from an RGB-NeRF or real images)
    enables training the $Z$-space NeRF and the AE.
    \textit{Right: inference approach}.
    After freezing the latent field and decoder, one can render the scene from an arbitrary viewpoint, obtaining a latent feature map that can be decoded into a high-resolution image. %
   }
   \label{fig:training.loop}
\end{figure*}

\section{Methods}
\label{sec:methods}

As in the standard NeRF scenario, 
    we expect only a set of multiview posed images,
    $ S_I = \{ (I_i, \Pi_i) \}_i $.
The goal is to learn a 3D scene representation in an autoencoder (AE) latent space, capable of novel view synthesis.
Thus, our model includes two neural modules 
    (\S\ref{sec:methods:arch}): 
    (i) a modified NeRF, $f$, 
        which outputs a latent vector 
        (in addition to its standard outputs),
    and
    (ii) an AE, 
        with encoder and decoder networks, $E$ and $D$.
To fit the model, we apply a multi-stage process: 
    training the AE, fitting the NeRF, and then fine-tuning $D$
    (see \S\ref{sec:methods:fitting}).

\subsection{ReLS-NeRF Neural Architecture}
\label{sec:methods:arch}
    
We first extend the standard colour-density field of NeRF 
    to include a latent feature vector, $z$, via
    $ f(x,r) = (\sigma\in\mathbb{R}_+, c\in[0,1]^3, z\in\mathbb{R}^n) $,
    where $x$ and $r$ represent the input position and direction, 
    and $\sigma$ and $c$ represent the output density and colour.
We refer to the $\sigma$ and $c$ fields as an ``RGB-NeRF'', 
    to distinguish them from the latent component of the ReLS-NeRF.
Note that the RGB-NeRF is used only in training,
    to learn the density field and produce renders to help train the latent component 
    (see \S\ref{sec:methods:fitting}).
Volume rendering is unchanged: 
    for a single feature at a pixel position, $p$, we use
\begin{equation}
    Z(p) = \int_{t_\mathrm{min}}^{t_\mathrm{max}} 
                \mathcal{T}(t) \sigma(t) z(t)\, dt,
\end{equation} 
to obtain the feature value at $p$, 
where $\mathcal{T}(t)$ is the transmittance \cite{tagliasacchi2022volume},
and $z(t)=z(x(t),r(t))$ is obtained by sampling the ray defined by $p$.
For camera parameters $\Pi$, 
    we denote the latent image rendering function as 
    $ \mathcal{R}(\Pi|f) = I_Z(\Pi) $, where $I_Z[p] = Z(p)$.
Replacing $z(t)$ with $c(t)$, for instance, would render colour in the standard manner, giving a colour image, $ I_C(\Pi) $ 
    (that does \textit{not} use $z$).
To obtain a colour image from $I_Z$,
    we simply pass it to the decoder, $D$; 
    i.e., view synthesis is 
    $ \widehat{I}_C(\Pi) = D(I_Z(\Pi)) $,
    which can be viewed as a form of 
    \textit{neural rendering}
    (e.g., \cite{niemeyer2021giraffe,tewari2020state,eslami2018neural}).
The benefit of using $\whi_C$ is that significantly fewer pixels need to be rendered, %
compared to $I_C(\Pi)$;
it also enables placing a prior on $\whi_C$ by choosing $D$ appropriately. 

We considered two choices of AE:
(i) the \textit{pretrained} VAE from Stable Diffusion~\cite{stable.diffusion}, 
    which we denote SD-VAE, 
and (ii) a smaller residual block-based AE 
    \cite{he2016deep,huang2018introvae} 
    (R32, when using a 32D latent space)
    that is randomly initialized.  
Both encoders provide an $8\times$ downsampling of the image. 

\subsection{Fitting Process}
\label{sec:methods:fitting}

A ReLS-NeRF is optimized in three stages: 
(A) AE training,
(B) joint NeRF fitting, and
(C) decoder fine-tuning.

\noindent
\textbf{AE training (A).} 
The first phase simply trains (or fine-tunes) the AE to reconstruct the training images of the scenes, using the mean-squared error.

\noindent
\textbf{Joint NeRF fitting (B).} 
In the second phase, we train the RGB and Latent components of the NeRF in conjunction with the decoder, $D$.
Our total loss function,
\begin{equation}\label{eq:ltotal}
    \mathfrak{L}_B = 
    \mathcal{L}_r + %
    \lambda_d \mathcal{L}_d + %
    \lambda_\mathrm{gr} \mathcal{L}_\mathrm{gr} +
    \mathcal{L}_p, %
\end{equation}
consists of 
the standard RGB loss on random rays, 
    $\mathcal{L}_r$,
the DS-NeRF~\cite{ds.nerf} depth loss, 
    $\mathcal{L}_d$,
the geometry regularizing distortion loss~\cite{mipnerf.360},
    $\mathcal{L}_\mathrm{gr}$,
and a patch-based loss for training the latent component,
    $\mathcal{L}_p$.
Given a posed image, $(I,\Pi)$,
    the latter loss is simply the error between 
    a sample patch, $\mathcal{P}\sim I$, and
    the corresponding rendered then decoded patch,
\begin{equation}
    \mathcal{L}_p = 
    \mathbb{E}_{\mathcal{P}\sim I, (I,\Pi)\sim S_I}
    \mathrm{MSE}( \mathcal{P}, D(I_Z(\Pi)) ).
\end{equation}

\noindent
\textbf{Decoder fine-tuning (C).} 
Finally, we fine-tune $D$, 
    utilizing a combination of the multiview posed images, $S_I$, and 
    renders from the RGB component of the ReLS-NeRF.
First, we sample random renders, 
    $\widetilde{S}_I = \{ (I_C(\Pi_s),\Pi_s) \,|\, \Pi_s\sim \Gamma(S_\Pi) \}_s$,
    where $ \Gamma(S_\Pi) $ is the uniform distribution over camera extrinsics, 
    obtained by interpolating between any triplet in $S_\Pi$.
Optimizing %
\begin{equation}
    \mathfrak{L}_C = 
        \gamma 
        \delta(S_I)
        + 
        (1 - \gamma) 
        \delta(\widetilde{S}_I),
\end{equation} 
where 
$ \delta(S) = \mathbb{E}_{(I,\Pi)\sim S} \mathrm{MSE}(I, \widehat{I}_C(\Pi)) $ and $\gamma\in[0,1]$ is a weighting hyper-parameter,
distills information from the RGB-NeRF into latent renderer.
See Fig.~\ref{fig:training.loop}.
Note that the real training images, $S_I$, are used;
    hence, the RGB-NeRF is \textit{not} strictly a ceiling on performance 
    (further, the presence of $D$ implies different generalization properties).

\subsection{Implementation Details}
We utilize the neural graphics primitives \cite{instant.ngp} architecture, 
    via the \verb|tiny-cuda-nn| library \cite{tiny.cuda.nn}.
All phases use Adam~\cite{adam} for optimization.
We remark that the loss gradient from the latent component 
    of the NeRF (i.e., from $\mathcal{L}_p$)
    is not back-propagated to the colour, $c$, and density, $\sigma$, fields.
Further, we use separate features for the latent feature vector, $z$, and $c$, 
    but render with the same $\sigma$.
In other words, RGB-NeRF training is unaffected by $z$.
For additional details, 
we refer the reader to our appendix.

\section{Experiments}

\subsection{Evaluation Metrics}

\subsubsection{Pixelwise and perceptual distances}
We measure performance with novel view synthesis on held-out test views.
In addition to the standard pixelwise peak signal-to-noise ratio (PSNR), we use perceptual losses to measure similarity, 
including LPIPS~\cite{perceptual}
and DreamSim~\cite{fu2023learning}.
LPIPS provides more human-like responses to low-level distortions
(e.g., noise, small colour/spatial shifts), %
whereas DreamSim is designed to be ``mid-level'' metric,
better capturing large-scale and semantic differences
than LPIPS 
(without being as high-level as, e.g., 
CLIP-based metrics 
\cite{radford2021learning,chan2022learning,vinker2022clipascene}).

\subsubsection{Local consistency}\label{sec:methods:localconsis}
When examining generative models of NeRFs that use decoders,
    we can qualitatively see 
    a ``shimmering'' effect in time
        (e.g., \cite{niemeyer2021giraffe,gu2021stylenerf}),
    which is also reminiscent 
        of generative video model artifacts
        (e.g., \cite{ho2022imagen,he2022latent}).
This jittering appears related to local appearance inconsistencies: 
    since each latent pixel corresponds to an RGB \textit{patch}.
    As $\Pi$ changes,
    interpolating in $z$-space 
    does not perfectly approximate the correct appearance changes.
This behaviour is distinct from the artifacts observed in standard NeRFs and we devise a simple metric to detect it: the
    {\textit{Reprojective Colour Consistency (RCC) metric}}.
The RCC measures sudden changes in appearance as $\Pi$ changes, 
    relying on the NeRF geometry to obtain correspondences.
Specifically, 
we reproject one render, $I_{i}$, 
into the reference frame of another, $I_{i+1}$,
using the NeRF depth, $D_i$, 
so
\begin{equation}
    \mathrm{RCC} = \mathrm{PSNR}\left( 
        \mathbb{E}_{i}[
        \mathrm{MSE}(
            I_{i+1}, \mathrm{Reproj}_{D_i,\Pi_{i+1}}{I_{i}}
        )]
    \right),
\end{equation}
where $I_i$ and $I_{i+1}$ are adjacent video frames. %
Notice that occlusions and view-dependent lighting effects will confound the RCC; however, these effects will (i) be relatively minimal across adjacent frames and (ii) be shared for the same scene, enabling it to be a fair comparative metric.

\begin{figure*}[t]
  \centering
   \includegraphics[width=0.99\linewidth]{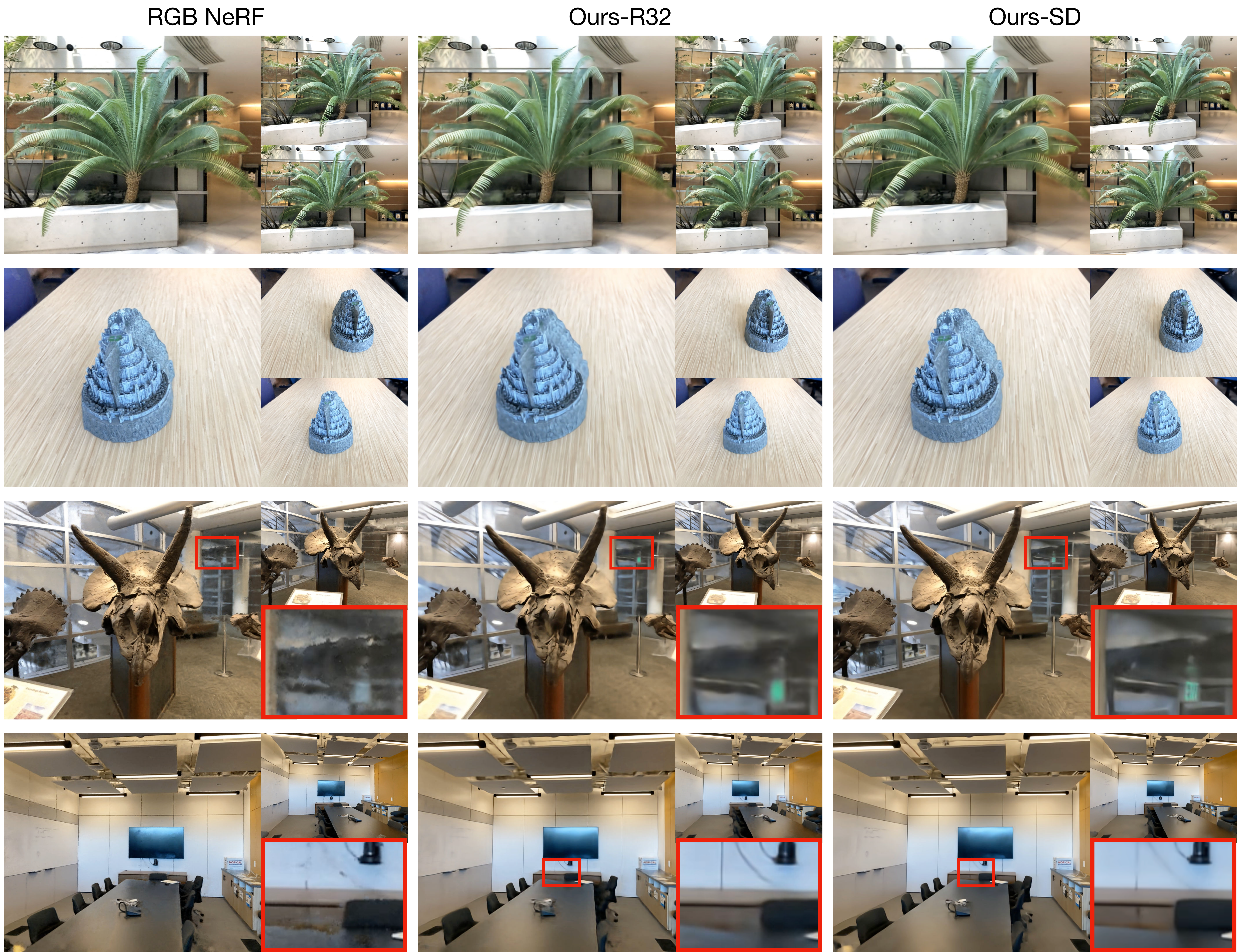}
   \caption{
        Qualitative comparison of NeRF renders.
        In the zoomed insets, we show how ReLS-NeRF-SD fixes some of the artifacts of the RGB-NeRF, despite being trained in part on its renders (see \S\ref{sec:methods:fitting}, phase C)
        One can also see the slight blur incurred by using the faster R32 AE (middle column).
        Notice that improvement in visual quality can actually have significant \textit{semantic} effects, in a manner that could impact downstream tasks (e.g., navigation): in the third row, for instance, one can actually read the ``exit'' sign in ReLS-NeRF-SD, but not in the other two cases.
   }
   \label{fig:qualitative.comparision}
\end{figure*}

\subsubsection{Video quality}
As noted above,
    adding a temporal dimension
    can make certain artifacts more perceptually detectable.
We therefore applied a recent video quality metric, 
    DOVER~\cite{wu2023dover}, 
    to NeRF-rendered videos.
DOVER has two components: 
    DOVER-aesthetic (DoA), 
        which focuses on high-level semantics,
    and
    DOVER-technical (DoT), 
        which detects low-level distortions 
        (e.g., blur and noise).
DOVER and the RCC are applied to 120-frame ``spiral video'' renders from the NeRF (as in LLFF~\cite{llff}).

{
\setlength{\tabcolsep}{3pt}
\begin{table}[t] %
    \centering
    \begin{tabular}{c|cccccc}
          & \multicolumn{3}{c}{Reference-based} & \multicolumn{3}{c}{Reference-free} \\
    NeRF  & PSNR$\uparrow$ & LPIPS$\downarrow$ & DS$\downarrow$ & DoA$\uparrow$ & DoT$\uparrow$ & RCC$\uparrow$ \\\hline
    RGB & 23.52 & 0.37 & \textbf{1.18} & 80.2 & 72.9 & \textbf{25.6} \\
    Ours-SD & \textbf{23.81} & \textbf{0.35} & 1.44 & \textbf{81.5} & \textbf{77.3} & 25.5 \\
    Ours-R32 & 23.37 & 0.40 & 1.71 & 76.4 & 74.3 & 25.3 \\
    \end{tabular}
    \caption{
        Test-view evaluation on eight LLFF scenes~\cite{llff}.
        Reference-based metrics %
        include PSNR, LPIPS \cite{perceptual}, and DreamSim (DS; $\times 10$) \cite{fu2023learning}.
        For reference-free metrics, 
            we use DOVER-technical (DoT), DOVER-aesthetic (DoA), 
            and our reprojective colour consistency (RCC) measure,
            computed on rendered videos. 
        Rows correspond to the standard RGB NeRF, the SDVAE-based ReLS-NeRF, and the R32-based ReLS-NeRF.
        ReLS-NeRF-SDVAE outperforms the RGB-space NeRF on the lower-level reference-based (PSNR and LPIPS) and reference-free (DoT) metrics,
            but has mixed performance 
            on the more semantic metrics (DS and DoA).
        Our RCC metric,  
            designed to detect the %
            ``shimmer'' present in decoded (neural rendered) videos, 
             detects slightly more inconsistency with ReLS-NeRF.
        Using R32 reduces accuracy, 
            but enables much faster rendering time 
            (see Table~\ref{tab:times}).
    }
    \label{tab:mets}
\end{table}
}

{ %
\begin{table}[t] %
    \centering
    \begin{tabular}{c|cccc}
    \multirowcell{2}{NeRF}      & \multirowcell{2}{Rendering Time} & \multicolumn{3}{c}{Fitting Time} \\
      $\,$  &  & (A) & (B) & (C) \\\hline
    RGB      & 132.1s [1$\times$] & \textbf{--} & \textbf{1h} & \textbf{--} \\
    Ours-SD  & 43.1s [3$\times$] & 10m & 2h & 2.5h \\
    Ours-R32 & \textbf{10.2s [13$\times$]} & 40m & 1.5h & 1.5h \\
    \end{tabular}
    \caption{
        Timings for inference (rendering 120 frames) 
            and fitting. %
        Changing the decoder architecture, $D$, trades off between efficiency and image quality.
        We measure the RGB-NeRF rendering time without the latent component.
    }
    \label{tab:times}
\end{table}
}

\subsection{Reconstruction Quality and Timing}
We display our evaluation in Table \ref{tab:mets}, 
    as well as timing measurements in Table \ref{tab:times},
    using eight LLFF scenes~\cite{llff} (see also Fig.~\ref{fig:qualitative.comparision} for qualitative examples)\protect\footnote{Images in Figs.~1-4 available in \href{https://drive.google.com/drive/folders/1M-_Fdn4ajDa0CS8-iqejv0fQQeuonpKF}{LLFF}~\cite{llff} under a \href{https://creativecommons.org/licenses/by/3.0}{CC BY 3.0 License}.},
    at $1008{\times}756$ resolution.
We see that ReLS-NeRF 
    (i.e., decoding a rendered latent feature map) with the
    SDVAE actually has superior novel view image quality, 
    while having superior inference speed (three times faster).
In particular, the low-level metrics, 
    including PSNR, LPIPS, and DoT, all
    prefer ReLS-NeRF-SD over the standard colour NeRF.
This is likely due to the fine-tuned decoder fixing artifacts 
    incurred by the colour NeRF, as can be seen
    in Fig.~\ref{fig:qualitative.comparision}.
The higher-level, more semantic metrics are more mixed:
    DreamSim prefers the RGB-NeRF, 
    while DoA slightly favours ReLS-NeRF-SD.

Among reference-based metrics, the semantically-oriented DreamSim is the only one by which the RGB-NeRF outperforms ReLS-NeRF-SD. 
Since DreamSim is a single-image metric, it is insensitive to temporal artifacts; however, DreamSim is known to be more sensitive to foreground objects \cite{fu2023learning}.
Interestingly, we qualitatively observe that ReLS-NeRF tends to improve image quality the most in scene areas far from the camera, where geometry is generally poorer quality -- i.e., in the background (see Fig.~\ref{fig:qualitative.comparision}).
Thus, one might speculate that such improvements are simply going unnoticed for DreamSim, which tends to focus on foreground objects of greater semantic importance.
    
\begin{figure*}[th]
\centering
\includegraphics[width=1.0\linewidth]{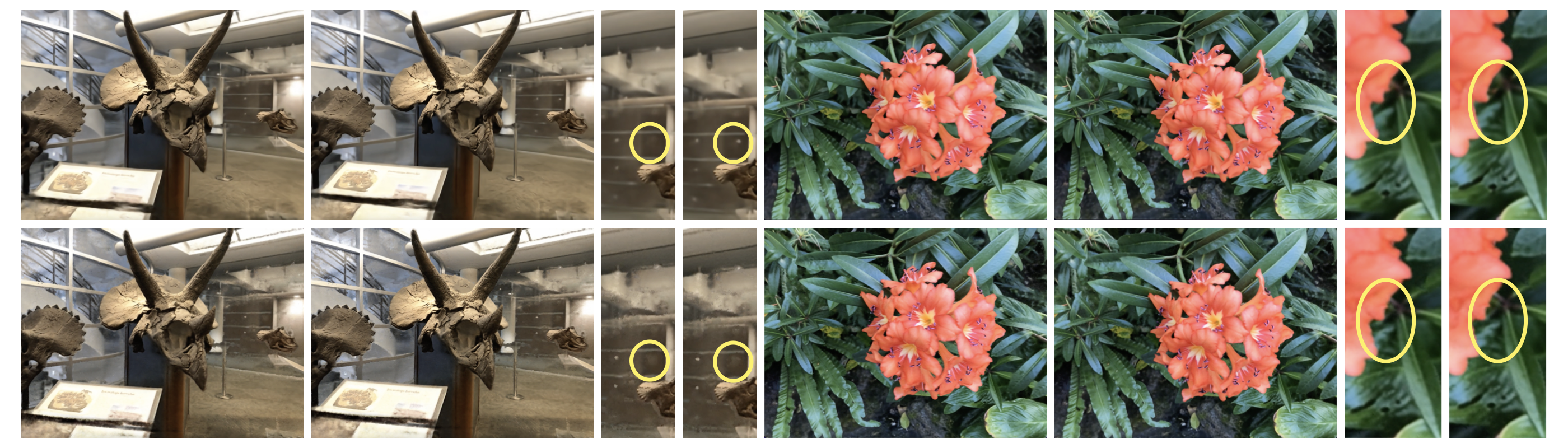}
\caption{
Examples of adjacent video frames from %
ReLS-NeRF-SD (top row) and the RGB NeRF (bottom row).
Each pair of images are temporally adjacent renders from a video. 
Notice, as in Fig.~\ref{fig:qualitative.comparision}, that ReLS-NeRF-SD has better per-frame image quality, as measured in the quantitative results of Table~\ref{tab:mets}.
For example, see the upper half of the leftward zoomed insets, where the RGB NeRF is more ``cloudy''.
However, there are \textit{temporal} artifacts that cannot be detected in a single frame (i.e., small cross-view appearance inconsistencies).
For instance, as can be seem in the highlighted areas of the zoomed insets, small spots can suddenly appear (left), or the shape of highlights can change (right).
This does not occur in the RGB NeRF, as volume rendering colours directly encourages view consistency, whereas the learned decoder in ReLS-NERF can introduce inconsistent appearances. 
This showcases the utility and need for our new reprojective colour consistency (RCC) metric (see \S\ref{sec:methods:localconsis}), which can capture these temporal aspects more directly.
}
\label{fig:frame.compare}
\end{figure*}

In addition, we find that the RCC prefers the RGB-NeRF over ReLS-NeRF.
Though it is hard to see in still images,
    ReLS-NeRF has slight temporal ``jittering'' artifacts,
    which the RCC is designed to detect.
We remark that other algorithms show similar view-inconsistencies across close frames (e.g., 3D generative models \cite{niemeyer2021giraffe} or video generators \cite{ho2022imagen}), and could potentially benefit from RCC estimates.
We illustrate this phenomenon with some examples in Fig.~\ref{fig:frame.compare}.
Due to the learned decoder, 
    unexpected appearance changes can occur across viewpoints. 
However, per-frame metrics, such as the traditionally applied LPIPS and PSNR, do not capture such inconsistencies; hence, ReLS-NeRF outperforms the RGB-NeRF on them (Table~\ref{tab:mets}). 
Interestingly, even the video metrics (DoT and DoA) prefer ReLS-NeRF, suggesting such algorithms are more sensitive to the cloudiness and noise artifacts of the standard NeRF, compared to the small jitters incurred by the neural rendering process.
In other words, by most metrics of quality (including the primary standard ones, PSNR and LPIPS), ReLS-NeRF is superior.

Finally, we show that the trade-off between 
    rendering efficiency and image quality 
    can be controlled by changing the AE architecture. 
Using R32 reduces inference time by {$\sim$}92\%,
    while decreasing test-view PSNR by only 0.15,
    compared to the RGB-NeRF rendering process.
In contrast to ReLS-NeRF-SD, while ReLS-NeRF-R32 does sacrifice some image quality 
    (e.g., ${\sim}$0.4 PSNR loss), it also reduces inference time by
    ${\sim}$76\%.
One can imagine choosing an architecture with the right level of trade-off for a given task.

\subsection{Ablations}
We find that removing phase C is devastating to ReLS-NeRF, 
    causing PSNR to drop to 22.85 (SD) and 20.87 (R32).
Since the SDVAE is pretrained, 
    ablating phase A has little effect on ReLS-NeRF-SD;
    however, doing so for ReLS-NeRF-R32 reduces
    PSNR by 0.1.
Note that the latter case trains the decoder, $D$, alongside the NeRF and then alone, in phases B and C.

\section{Discussion}

We have shown that ReLS-NeRF can improve image quality, %
    while being several times faster to render. 
In  particular, the SD-based ReLS-NERF is superior on the main metrics commonly used to evaluate NeRFs on test views (i.e., PSNR and LPIPS), as well as on a state-of-the-art reference-free video quality estimator. 
Empirically, we observed that current image and video evaluation metrics do not obviously capture temporal artifacts that are characteristic of ReLS-NeRF, caused by view-inconsistent appearance changes (due to the learned component within the rendering process).
Hence, we introduced a simple metric for detecting such anomalies. 
Further, we have demonstrated a tradeoff between efficiency and quality,
    which can be controlled by the architecture of the AE.
Importantly, to obtain its speedup,
    ReLS-NeRF does not ``bake'' the scene or transform to a mesh;
    hence, e.g., it could still be continually
    trained online in the standard fashion.
In other words, it retains a number of useful properties of standard NeRFs (e.g., differentiability and access to an implicit 3D shape field), while gaining additional efficiency and image quality.

For many robotics tasks, \textit{fast differentiable rendering} is a key component for online learning of 3D scene representations. 
This includes simultaneous localization and mapping, navigation, and modelling the dynamics of the environment (i.e., ensuring the internal representation is up-to-date, given perceptual inputs).
We feel that ReLS-NeRF is well-suited for such situations, 
    as it retains differentiability, while improving rendering efficiency and even image quality as well.
Other promising future directions include
    utilizing different AEs to provide task-specific biases 
        (e.g., for 3D scene editing, faster speed, or higher image quality),
    improving the AE architecture to suit this scenario
        (e.g., devising a geometry-aware decoder),
    and better customizing the volume rendering process to latent space rendering (e.g., using a learned mapping instead of volume integration).

\section*{APPENDIX}

\subsection{Additional Implementation Details}
When training, we used 
    $\lambda_d = 0.1$,
    $ \gamma = 0.7$, and 
    $\lambda_\mathrm{gr} = 10^{-3} / 2$.
The NeRF architecture was the same as previous works based on Instant-NGP (see \cite{spinnerf}).
The LLFF scenes used were 
\verb|fern|, \verb|horns|, \verb|orchids|, \verb|flower|,
\verb|leaves|, \verb|room_tv|, \verb|trex|, and \verb|fortress|.

\subsection{Fitting Hyper-Parameters}
\textbf{Phase A.}
The SDVAE/R32 NeRFs were optimized for 500/3000 iterations,
    using learning rates of $10^{-4}$/$4\times 10^{-4}$.
The learning rates were halved at 150, 300, and 450 iterations (SDVAE) and every 500 iterations for R32.
Patches of size $512^2$ were used, with batch sizes of 3/5.

\textbf{Phase B.}
The joint optimization was run for 20K iterations.
We used 4096 rays for the colour and DS-NeRF losses, each.
The latent loss, $\mathcal{L}_p$, is computed via $32^2$ latent-space patches.
The learning rate (excluding the VAE) 
starts from $10^{-2}$ and 
is decayed according to 
$10^{-2} \times (10^{-1})^{ t / \tau }$,
where $t$ is the step iteration and $\tau=10^4$.
The VAE is optimized with a fixed learning rate of $10^{-4}$.

\textbf{Phase C.}
Decoder fine-tuning proceeds for 3000/10000 iterations for the SDVAE/R32 architecture.
A batch size of three was used (one from $S_I$ and two from $\widetilde{S}_I$).
Note that we render 512 images from the RGB-NeRF to act as supervision (i.e., $|\widetilde{S}_I| = 512$).
The process starts from a learning rate of $10^{-4}$, 
    and is decayed by 0.5 every 1000/2500 iterations.

\subsection{R32 Architecture}
The encoder, $E$, has the following structure:
\verb|C5|, \verb|RBIN|, \verb|HD|, \verb|RBIN|, \verb|HD|, \verb|RBIN|, \verb|HD|, \verb|RBIN|, \verb|C1|.
The components are as follows:
\verb|C5| is a conv-$5{\times}5$-\verb|norm|-elu block;
\verb|RBIN| is two residual blocks~\cite{he2016deep}, 
    each using conv-$3{\times}3$ and \verb|norm|;
\verb|HD| is a bilinear halving downscaler; and
\verb|C1| is just a conv-$1{\times}1$. 
The encoder has layer sizes of
\verb|(32,128,128,256,256)|.

The decoder, $D$, has the following structure:
\verb|C1|, \verb|RBIN|, \verb|HU|, \verb|RBIN|, \verb|HU|, \verb|RBIN|, \verb|HU|, \verb|RBIN|, \verb|C1|, \verb|sigmoid|.
Components are the same, 
    except that \verb|HU| is a bilinear doubling upscaler.
The decoder has layer sizes of
\verb|(256,256,128,128,32)|.

Both networks use the 
ELU non-linearity~\cite{clevert2015fast} and
instance normalization~\cite{ulyanov2016instance} 
as \verb|norm|.

\clearpage

\bibliographystyle{IEEEtran}
\bibliography{IEEEabrv,defs,egbib}

\end{document}